\pdfoutput=1
\documentclass{article}

\usepackage[preprint]{neurips_2026}

\usepackage[utf8]{inputenc}
\usepackage[T1]{fontenc}
\usepackage{hyperref}
\usepackage{url}
\usepackage{booktabs}
\usepackage{amsfonts,amsmath,amssymb}
\usepackage{nicefrac}
\usepackage{microtype}
\usepackage{xcolor}
\usepackage{graphicx}
\usepackage{tikz}
\usetikzlibrary{arrows.meta,positioning,fit,calc}
\usepackage{enumitem}
\usepackage{makecell}
\usepackage{float}

\hypersetup{colorlinks=true,linkcolor=blue!50!black,citecolor=blue!50!black,urlcolor=blue!50!black,
  pdftitle={ClockRace: Temporal Graph Prediction from Addressed Time-Referenced Records},pdfauthor={Tero Keski-Valkama}}

\newcommand{\codeurl}{\url{https://github.com/Hypersleep-ai/clockrace-tgb}}

\newcommand{\mainrows}{%
\texttt{tgbn-trade} & NAViS & 0.863 & 0.8746\,$\pm$\,0.0004 & \textbf{0.8687}\,$\pm$\,0.0013 & +0.0057 & 5/5 & 2,107 & 4 CPU \\
\texttt{tgbn-genre} & NAViS & 0.528 & 0.5387\,$\pm$\,0.0002 & \textbf{0.5465}\,$\pm$\,0.0002 & +0.0185 & 5/5 & 2,360 & 10 CPU \\
\texttt{tgbn-reddit} & NAViS & 0.569 & 0.6008\,$\pm$\,0.0002 & \textbf{0.5868}\,$\pm$\,0.0002 & +0.0178 & 5/5 & 2,360 & 443 CPU \\
\texttt{tgbn-token} & NAViS & 0.513 & 0.5022\,$\pm$\,0.0004 & \textbf{0.5156}\,$\pm$\,0.0006 & +0.0026 & 5/5 & 2,039 & 350 CPU \\
\texttt{tgbl-wiki-v2} & TRACE & 0.833 & 0.8518\,$\pm$\,0.0001 & \textbf{0.8353}\,$\pm$\,0.0002 & +0.0023 & 5/5 & 7,995 & 35 CPU \\
\texttt{tgbl-review-v2} & GraphMixer & 0.521 & 0.4883\,$\pm$\,0.0004 & \textbf{0.5398}\,$\pm$\,0.0008 & +0.0188 & 5/5 & 11,642,367 & 34 CPU \\
\texttt{tgbl-flight-v2} & TPNet & 0.884 & 0.9147\,$\pm$\,0.0001 & \textbf{0.9085}\,$\pm$\,0.0002 & +0.0245 & 5/5 & 6,929 & 148 CPU \\
\texttt{tgbl-coin-v2} & TPNet & 0.832 & 0.8536\,$\pm$\,0.0002 & \textbf{0.8691}\,$\pm$\,0.0006 & +0.0371 & 5/5 & 12,737 & 313 CPU \\
\texttt{thgl-github} & EdgeBank$_\infty$ & 0.413 & 0.7558\,$\pm$\,0.0003 & \textbf{0.8466}\,$\pm$\,0.0007 & +0.4336 & 5/5 & 7,157 & 70 CPU \\
\texttt{thgl-myket} & EdgeBank$_\infty$ & 0.456 & 0.9166\,$\pm$\,0.0001 & \textbf{0.9158}\,$\pm$\,0.0007 & +0.4598 & 5/5 & 7,061 & 198 CPU \\
\texttt{thgl-forum} & TGN, edge types & 0.729 & 0.8167\,$\pm$\,0.0003 & \textbf{0.8162}\,$\pm$\,0.0002 & +0.0872 & 5/5 & 8,051 & 821 CPU \\
\texttt{tkgl-polecat} & CountTRuCoLa & 0.256 & 0.2633\,$\pm$\,0.0001 & \textbf{0.2581}\,$\pm$\,0.0002 & +0.0021 & 5/5 & 29,081 & 89 CPU \\
\texttt{tkgl-icews} & CountTRuCoLa & 0.214 & 0.3470\,$\pm$\,0.0001 & \textbf{0.3452}\,$\pm$\,0.0004 & +0.1312 & 5/5 & 34,438 & 513 CPU \\
\texttt{tkgl-wikidata} & CountTRuCoLa & 0.609 & 0.7408\,$\pm$\,0.0001 & \textbf{0.6707}\,$\pm$\,0.0002 & +0.0617 & 5/5 & 81,178 & 18 GPU \\}
\newcommand{\seedrows}{%
\texttt{tgbn-trade} & 0.8747/0.8683 & 0.8740/0.8674 & 0.8745/0.8683 & 0.8745/0.8708 & 0.8751/0.8686 \\
\texttt{tgbn-genre} & 0.5385/0.5467 & 0.5388/0.5462 & 0.5387/0.5466 & 0.5387/0.5464 & 0.5390/0.5467 \\
\texttt{tgbn-reddit} & 0.6010/0.5869 & 0.6008/0.5868 & 0.6008/0.5867 & 0.6010/0.5870 & 0.6005/0.5866 \\
\texttt{tgbn-token} & 0.5022/0.5156 & 0.5019/0.5153 & 0.5029/0.5162 & 0.5021/0.5162 & 0.5019/0.5149 \\
\texttt{tgbl-wiki-v2} & 0.8518/0.8353 & 0.8519/0.8350 & 0.8517/0.8356 & 0.8519/0.8354 & 0.8517/0.8352 \\
\texttt{tgbl-review-v2} & 0.4880/0.5393 & 0.4886/0.5405 & 0.4877/0.5387 & 0.4882/0.5400 & 0.4887/0.5408 \\
\texttt{tgbl-flight-v2} & 0.9148/0.9086 & 0.9146/0.9083 & 0.9148/0.9087 & 0.9147/0.9087 & 0.9146/0.9085 \\
\texttt{tgbl-coin-v2} & 0.8537/0.8694 & 0.8536/0.8690 & 0.8538/0.8698 & 0.8533/0.8689 & 0.8534/0.8682 \\
\texttt{thgl-github} & 0.7558/0.8472 & 0.7562/0.8475 & 0.7555/0.8458 & 0.7559/0.8461 & 0.7555/0.8464 \\
\texttt{thgl-myket} & 0.9165/0.9145 & 0.9167/0.9161 & 0.9166/0.9158 & 0.9166/0.9161 & 0.9166/0.9164 \\
\texttt{thgl-forum} & 0.8171/0.8164 & 0.8166/0.8161 & 0.8164/0.8161 & 0.8168/0.8164 & 0.8165/0.8161 \\
\texttt{tkgl-polecat} & 0.2635/0.2584 & 0.2632/0.2581 & 0.2632/0.2579 & 0.2632/0.2580 & 0.2634/0.2581 \\
\texttt{tkgl-icews} & 0.3472/0.3454 & 0.3470/0.3457 & 0.3470/0.3447 & 0.3469/0.3450 & 0.3470/0.3454 \\
\texttt{tkgl-wikidata} & 0.7406/0.6703 & 0.7409/0.6708 & 0.7409/0.6705 & 0.7408/0.6708 & 0.7409/0.6709 \\}
\newcommand{\earlierrows}{%
\texttt{tgbn-genre} & single decay-scale bank (v2) & 0.5365\,$\pm$\,0.0002 & 0.5441\,$\pm$\,0.0002 & 5/5 & 0.5465 \\
\texttt{tgbn-reddit} & single decay-scale bank (v2) & 0.5981\,$\pm$\,0.0002 & 0.5840\,$\pm$\,0.0002 & 5/5 & 0.5868 \\
\texttt{tkgl-polecat} & without the two-hop path cause & 0.2596\,$\pm$\,0.0001 & 0.2544\,$\pm$\,0.0002 & 0/5 & 0.2581 \\}
\newcommand{\costrows}{%
\texttt{tgbn-trade} & 255 & -- & 2,107 & -- & 4 & AB \\
\texttt{tgbn-genre} & 992 + 513 & -- & 2,360 & 3.6 & 10 & A \\
\texttt{tgbn-reddit} & 11,068 + 698 & -- & 2,360 & 6.7 & 443 & A \\
\texttt{tgbn-token} & 60,755 + 1,001 & -- & 2,039 & 11.9 & 350 & A \\
\texttt{tgbl-wiki-v2} & 9,227 & -- & 7,995 & -- & 35 & AB \\
\texttt{tgbl-review-v2} & 352,637 & 728,919 & 11,642,367 & 3.2 & 34 & A \\
\texttt{tgbl-flight-v2} & 18,143 & 10,026,943 & 6,929 & 16.9 & 148 & A \\
\texttt{tgbl-coin-v2} & 638,486 & 3,421,411 & 12,737 & 9.8 & 313 & A \\
\texttt{thgl-github} & 5,856,765 & 2,624,932 & 7,157 & 6.2 & 70 & A \\
\texttt{thgl-myket} & 1,530,835 & 8,044,915 & 7,061 & 17.1 & 198 & A \\
\texttt{thgl-forum} & 152,816 & 3,563,653 & 8,051 & 12.1 & 821 & A \\
\texttt{tkgl-polecat} & 150,931 & 532,636 & 29,081 & 5.8 & 89 & C \\
\texttt{tkgl-icews} & 87,856 & 4,651,378 & 34,438 & 27.3 & 513 & A \\
\texttt{tkgl-wikidata} & 1,226,440 & 2,877,500 & 81,178 & 22.6 & 18 & C (GPU) \\}
\newcommand{\publishedrows}{%
\texttt{tgbn-trade} & 0.8687 & NAViS & 0.863\,$\pm$\,0.001 & $+0.0057$ & 5/5 & DyNA-Mixer 0.869$^{a}$ \\
\texttt{tgbn-genre} & 0.5465 & DyNA-Mixer & 0.544\,$\pm$\,0.000 & $+0.0025$ & 5/5 & -- \\
\texttt{tgbn-reddit} & 0.5868 & DyNA-Mixer & 0.590\,$\pm$\,0.000 & $-0.0032$ & 0/5 & -- \\
\texttt{tgbn-token} & 0.5156 & DyNA-Mixer & 0.523\,$\pm$\,0.001 & $-0.0074$ & 0/5 & -- \\
\texttt{tgbl-wiki-v2} & 0.8353 & TRACE$^{e}$ & 0.83288\,$\pm$\,0.00008 & $+0.0024$ & 5/5 & PRISM-TNCN 0.920$^{g}$ \\
\texttt{tgbl-review-v2} & 0.5398 & TSNN & 0.5400\,$\pm$\,0.0006 & $-0.0002$ & 2/5 & \makecell[lt]{TGR 0.854, 0.599$^{b}$;\\ A2QTGN 0.588$^{d}$} \\
\texttt{tgbl-flight-v2} & 0.9085 & CL-DGFM & 0.9068\,$\pm$\,0.0005 & $+0.0017$ & 5/5 & CRAFT-R 0.914$^{c}$ \\
\texttt{tgbl-coin-v2} & 0.8691 & CTDG-SSM$^{e}$ & 0.862\,$\pm$\,0.003 & $+0.0071$ & 5/5 & \makecell[lt]{CRAFT-R 0.885$^{c}$;\\ PRISM-TNCN 0.913$^{f,g}$} \\
\texttt{thgl-github} & 0.8466 & TSNN & 0.8336\,$\pm$\,0.0001 & $+0.0130$ & 5/5 & -- \\
\texttt{thgl-myket} & 0.9158 & TSNN & 0.9219\,$\pm$\,0.0009 & $-0.0061$ & 0/5 & -- \\
\texttt{thgl-forum} & 0.8162 & EdgeReMIND & 0.7306\,$\pm$\,0.0021 & $+0.0856$ & 5/5 & \makecell[lt]{TSNN 0.838$^{h}$;\\ RF-Logic 0.739$^{i}$} \\
\texttt{tkgl-polecat} & 0.2581 & CognTKE & 0.287 & $-0.0289$ & 0/5 & RF-Logic 0.424$^{i}$ \\
\texttt{tkgl-icews} & 0.3452 & TSNN & 0.2871\,$\pm$\,0.0003 & $+0.0581$ & 5/5 & -- \\
\texttt{tkgl-wikidata} & 0.6707 & TSNN & 0.6502\,$\pm$\,0.0022 & $+0.0205$ & 5/5 & -- \\}
\newcommand{\leakagefreerows}{%
\texttt{tgbn-genre}, leakage-free & 0.5330 & hybrid blend & 0.530 & $+0.0030$ & 5/5 & -- \\
\texttt{tgbn-reddit}, leakage-free & 0.5663 & hybrid blend & 0.561 & $+0.0053$ & 5/5 & -- \\
\texttt{tgbn-token}, leakage-free & 0.4580 & hybrid blend & 0.454 & $+0.0040$ & 5/5 & -- \\}
\newcommand{\leaderboardread}{24 November 2026}

\newcommand{\publishedleadcount}{7}
\newcommand{\publishedleadmin}{$+0.0017$ MRR on \texttt{tgbl-flight-v2}}
\newcommand{\publishedleadmax}{$+0.086$ MRR on \texttt{thgl-forum}}
\newcommand{\publishedtrails}{\texttt{thgl-myket}, \texttt{tkgl-polecat}}
\newcommand{\levelclause}{, and is level on \texttt{tgbl-review-v2}}
\newcommand{\unresolvedclause}{; on \texttt{tgbl-wiki-v2} and \texttt{tgbl-coin-v2}, where higher published scores come from test paths that read events of the batch being predicted, the comparisons are not resolved}
\newcommand{\unresolvednote}{ $^e$Strongest result evaluated without the reads marked $^f$ or $^g$; the comparison is not resolved and not counted.}
\newcommand{\publishedbarcites}{NAViS~\citep{feldman2026navis}, DyNA-Mixer~\citep{zheng2026dynamixer}, TRACE~\citep{wu2026trace}, CL-DGFM~\citep{ma2026cldgfm}, CTDG-SSM~\citep{raghuvanshi2026ctdgssm}, TSNN~\citep{anon2026tsnn}, EdgeReMIND~\citep{pollard2026edgeremind}, and CognTKE~\citep{chen2025cogntke} as run by \citet{gastinger2026counttrucola}}

\newcommand{\prismcaptionnote}{ $^f$Memory also reads earlier events of the query's own batch at its timestamp. $^g$The decoder's common-neighbour graph links the query's own rows to rows of later events of its batch (PRISM~\citep{islam2026prism}).}
\newcommand{\prismparagraph}{\textbf{Wiki and coin.} PRISM~\citep{islam2026prism} prints 0.9199 on \texttt{tgbl-wiki-v2} and 0.9130 on \texttt{tgbl-coin-v2} with TNCN (one run each, at the batch size with the highest test MRR; its Section 6.3 and Table 2), above our 0.8353 and 0.8691. It ranks each test edge against TGB's negatives, and its neighbour sampler reads strictly earlier timestamps, but the test path of its release also reads events of the batch being predicted, through two channels. First, its common-neighbour decoder works on one graph per batch whose sampled neighbours are replaced by version rows, the rows of the batch's latest events that touch them (\texttt{modules/train\_utils.py} lines 823--825 at commit \texttt{a7b940f}); these are source and true-destination rows of the batch's events, never a negative's row (\texttt{mem\_update\_graph.cu} lines 5--31), and the decoder symmetrises the edge list (\texttt{modules/NCNDecoder/NCNPred.py} lines 99--103), so a query's own rows are linked to rows of later events of its batch. Training builds the same graph, so the decoder learns to read it. On wiki validation (23,621 queries; PRISM's own functions with random features, nothing trained), this common-neighbour input is non-zero for 67.6\% of true destinations against 2.4\% of random negatives, and for 92.2\% against 5.1\% on pairs without an earlier event; without version rows it is zero. Second, its memory reads the earlier events of a query's own batch (\texttt{mem\_update\_graph.cu} lines 18--27), among them events at the query's own timestamp, which our reads never use (Section~\ref{sec:protocol}); on coin validation, where a timestamp holds 19.5 events on average, these touch the true destination of 20.1\% of queries, against 0.004\% for a random node. PRISM's authors revised their code on 2 December 2026 to read only strictly earlier events in the batch's version rows and in the common-neighbour decoder (commits \texttt{43e57c3} and \texttt{df943da}); revised scores have not been published. A paired check that we registered on 1 December trained PRISM-TNCN on wiki with the authors' code and configuration and read the official test set once in each arm: as released it scored 0.91496, and our implementation with strictly earlier reads, whose decoder reads raw sampled edges, scored 0.74335. The arms differ in more than the batch's events. On the first 32 queries of six validation batches, the authors' revised decoder gives a non-zero common-neighbour input for 57 of 192 true destinations where ours gives none, and none of its inputs changes when the batch's later events are removed. So 0.74335 measures our implementation, not PRISM, and the two test scores do not isolate the effect of the batch's events. PRISM's scores without these reads are unpublished, so both comparisons are not resolved: ClockRace exceeds TRACE's 0.8329 on wiki and CTDG-SSM's 0.862 on coin, the strongest results evaluated without them, with every seed, and is below PRISM's printed scores.}
\newcommand{\smallmarginsnote}{On polecat, and on token under TGB's official loop, results published outside the leaderboards are above ours; on wiki the comparison with a higher printed score is not resolved (Section~\ref{sec:published}).}
\newcommand{\cleanroomreads}{A clean-room reproduction from the public repository read one registered seed per dataset, at these UTC times: review (seed 3, 12:36), github (seed 4, 13:25), trade (seed 2, 13:33), polecat (seed 1, 17:15) and wiki (seed 0, 21:02) on 29 November; icews (seed 0, 00:06), forum (seed 1, 02:18), genre (seed 2, 02:41), coin (seed 1, 03:26), myket (seed 3, 05:41), flight (seed 0, 17:55) and reddit (seed 1, 21:20) on 30 November. Its token run from public commit \texttt{4182dbc} stopped at its first validation pass, before reading test, on 1 December at 02:38 UTC: that commit's download step left py-tgb's processing to the first load, which returns the raw token amounts (up to $1.16\times10^{77}$) rather than the log-normalized amounts that every reported run read, and the 32-bit affinity state overflowed. The same seed, rerun from public commit \texttt{d3ef3d9}, which processes the data during the download, read test at 08:01 UTC on 1 December. Every read lies inside its five-seed range, and trade, genre, wiki and polecat reproduce bit for bit. The wikidata run had not read by 3 December 2026, 10:46 UTC, and no other test read of a model of ours on these datasets was made by then.}
\newcommand{\reviewsentence}{We ran the authors' code for one training epoch and scored the first 40,000 validation queries both ways from the same state: the released rule reads 0.3162 MRR and a label-blind rule, in which every previously observed candidate takes the expander path, reads 0.2789 (difference $+0.0373$, 95\% bootstrap interval $+0.0358$ to $+0.0387$). Under the released rule the true destination takes the expander path in 44.1\% of queries and 0.12\% of negative candidates take it; under the label-blind rule 9.0\% do. TGR's label-blind scores after full training are unpublished. TSNN~\citep{anon2026tsnn}, evaluated with TGB's negatives and evaluator and without this dependence, prints $0.5400\pm0.0006$; its review run reads each earlier review's star rating, which is given with the review. ClockRace's 0.5398 is level with it, with two of five seeds above it, and is below TGR's printed scores; CL-DGFM's $0.5289\pm0.0020$~\citep{ma2026cldgfm} follows.}

\title{ClockRace: Temporal Graph Prediction from\\Addressed Time-Referenced Records}

\author{Tero Keski-Valkama\\
  Hypersleep (hypersleep.ai)\\
  \texttt{tero.keski-valkama@hypersleep.ai}}

\begin{document}

\maketitle

\begin{abstract}
Temporal graph benchmarks reward models that keep the right evidence about who interacted with whom, and when.
We present ClockRace, a family of temporal graph predictors whose state is a set of sparse records addressed by the
entities, pairs and relations that later queries name. Each record holds a few numerical history values and a time
reference. An interaction updates only the records it addresses, and a query brings the records of its source and its
candidates to the query time. Candidate-specific causes read these records, a small learned map turns each read into a
hazard, and the prediction is the race among the candidates: the probability that a candidate's clock fires first,
a softmax over log-hazards. On link, heterogeneous and knowledge-graph tasks, training draws each race's competitors from the
training data by the rule the benchmark uses to sample its negatives. On 14 datasets of TGB and TGB~2.0, covering node
affinity, link prediction,
and heterogeneous and knowledge-graph forecasting, the mean test score over five seeds exceeds the best posted
test score on each of these leaderboards, as read on \leaderboardread{}, and every seed exceeds it. Compared with the
strongest published results outside the leaderboards that use the same information, ClockRace leads on
\publishedleadcount{} of the 14 datasets with every seed, by \publishedleadmin{} to \publishedleadmax{}, and on
\texttt{tgbn-reddit} and \texttt{tgbn-token} under a leakage-free protocol; it trails on \publishedtrails{} and, under
TGB's official loop, on reddit and token\levelclause{}\unresolvedclause{}. The models use no per-node embeddings, train 2,039
to 81,178 parameters, and run on one CPU thread per seed, or one GPU for the 1.2-million-entity \texttt{tkgl-wikidata}.
\end{abstract}

\section{Introduction}

We have a hypothesis: much of the world is a set of hidden processes running in parallel and linked by causal
relations, and what we observe of it is an asynchronous stream of events. A model built for such a world is sparse,
asynchronous and causal, with parallel clocks that race, delays that carry information and a race that performs
computation~\citep{keskivalkama2021sleeping}. Such a model should have the right inductive bias for such data, and should spend work only where events
happen.
Temporal graph benchmarks test both consequences directly.

Temporal graphs record timestamped interactions between entities: users editing pages, accounts transferring tokens,
airports exchanging flights, countries trading goods, actors in political events. The Temporal Graph Benchmark
(TGB)~\citep{huang2023tgb} and TGB~2.0~\citep{gastinger2024tgb2} turned prediction on such streams into reproducible
leaderboards with chronological splits, fixed negative samples and official evaluators.

EdgeBank, a memory of seen edges~\citep{poursafaei2022edgebank}, is a strong reference on link prediction, and the
only posted entry on two TGB~2.0 heterogeneous graphs. A persistent or moving-average forecast~\citep{huang2023tgb} is
second on all four node-affinity leaderboards. A rule learner~\citep{gastinger2026counttrucola} leads three of the four
knowledge-graph leaderboards, and the recurrency baseline~\citep{gastinger2024recurrency} is second on
\texttt{tkgl-icews}. Recency and popularity
heuristics rival trained networks on link prediction~\citep{cornell2025heuristics,wu2026trace}. Meanwhile, most
neural methods evaluated on the larger TGB~2.0 graphs exceeded 40\,GB of GPU memory or a week of
runtime~\citep{gastinger2024tgb2}. The bottleneck in these tasks is how evidence is stored, addressed and read at the
moment of a query.

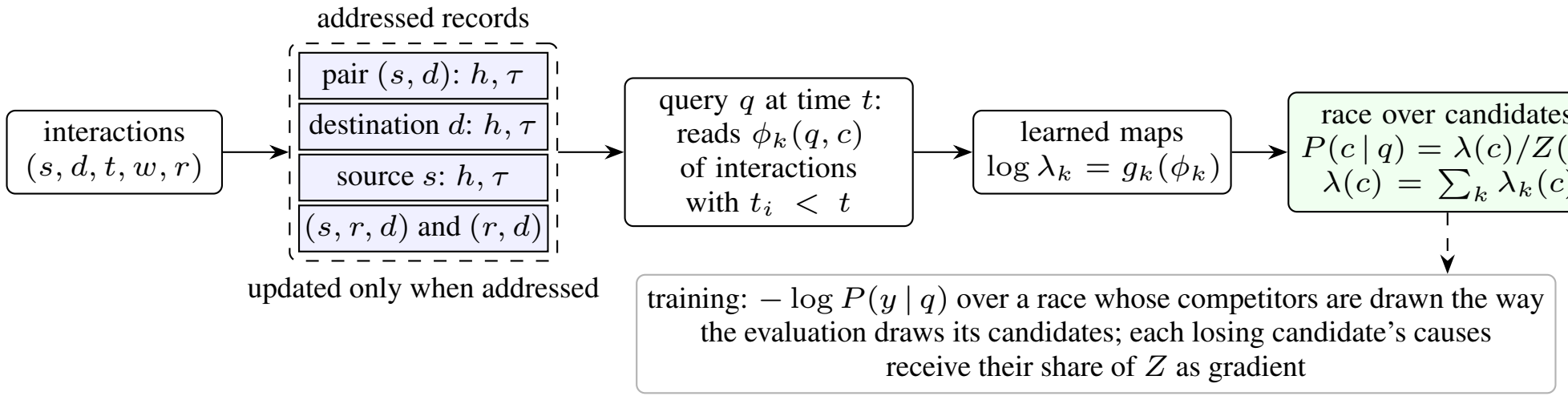
\begin{figure}[t]
\centering
\begin{tikzpicture}[font=\scriptsize,>={Stealth[length=1.6mm]},
  box/.style={draw,rounded corners=2pt,align=center,inner sep=2.5pt,fill=white},
  rec/.style={draw,minimum width=2.05cm,minimum height=0.38cm,inner sep=1pt,fill=blue!6},
  lab/.style={align=center}]
\node[box,text width=1.6cm] (stream) at (0.95,0) {interactions\\ $(s,d,t,w,r)$};
\node[rec] (r1) at (3.5,0.63) {pair $(s,d)$: $h,\tau$};
\node[rec] (r2) at (3.5,0.21) {destination $d$: $h,\tau$};
\node[rec] (r3) at (3.5,-0.21) {source $s$: $h,\tau$};
\node[rec] (r4) at (3.5,-0.63) {$(s,r,d)$ and $(r,d)$};
\node[draw,dashed,rounded corners=2pt,fit=(r1)(r2)(r3)(r4),inner sep=2pt] (mem) {};
\node[lab,above=0pt of mem] {addressed records};
\node[lab,below=0pt of mem] {updated only when addressed};
\draw[->] (stream.east) -- (mem.west);
\node[box,text width=2.2cm] (read) at (6.35,0) {query $q$ at time $t$:\\ reads $\phi_k(q,c)$\\ of interactions\\ with $t_i<t$};
\draw[->] (mem.east) -- (read.west);
\node[box,text width=1.95cm] (map) at (9.1,0) {learned maps\\ $\log\lambda_k=g_k(\phi_k)$};
\draw[->] (read.east) -- (map.west);
\node[box,fill=green!6,text width=2.45cm] (race) at (11.95,0) {race over candidates\\ $P(c\,|\,q)=\lambda(c)/Z(q)$\\ $\lambda(c)=\sum_k\lambda_k(c)$};
\draw[->] (map.east) -- (race.west);
\node[lab,draw=gray!60,rounded corners=2pt,inner sep=2.5pt] (train) at (9.05,-1.5) {training: $-\log P(y\,|\,q)$ over a race whose competitors are drawn the way\\ the evaluation draws its candidates; each losing candidate's causes\\ receive their share of $Z$ as gradient};
\draw[->,dashed] (race.south) -- (race.south |- train.north);
\end{tikzpicture}
\caption{ClockRace. Interactions update only the records they address (pair, destination, source, relation-addressed
families). A query reads the records of its source and of each candidate, transported to the query time, from
interactions strictly earlier than the query. Learned maps turn the reads into cause hazards, and the candidates race.}
\label{fig:method}
\end{figure}

ClockRace builds the predictor around that observation (Figure~\ref{fig:method}). State is held in \emph{records} addressed by what later queries
name: a source--destination pair, a relation and its destinations, an entity. A record holds a few numerical history
values (decayed counts, damped rotations) and the time at which they were last refreshed. An interaction
updates only the records it addresses, so the cost of an update does not grow with the graph. A query reads the records
of its source and of each candidate, transported to the query time. Because the model is causal, a query never reads an
interaction at or after its own time. A small learned map turns each read into a hazard, and the candidates race: the predicted probability of a
candidate is the probability that its exponential clock fires first.

\paragraph{Contributions.}
\begin{itemize}[leftmargin=1.2em,itemsep=1pt,topsep=2pt]
\item \textbf{A formulation} that covers node affinity, link prediction, and heterogeneous and knowledge-graph
forecasting with one state model (addressed time-referenced records) and one readout (a race among candidates over a sum
of cause hazards). On knowledge graphs the race is exact over all entities: sparse causes add hazard only to the
candidates they propose, over a dense background that has no per-entity parameters (Section~\ref{sec:method}).
\item \textbf{Evaluation-shaped training races}: competitors drawn the way the benchmark's negative generator draws
them. On \texttt{tgbl-review-v2}, where up to half of each query's negatives are the reviewer's own past items, this race
alone teaches the model that reviewers rarely return to an item (Section~\ref{sec:analysis}).
\item \textbf{Results on 14 datasets} under a fully disclosed protocol, compared with the posted leaderboards
(Section~\ref{sec:results}) and with the strongest published results outside them (Section~\ref{sec:published}). Each
configuration was registered before its first test read. Five seeds were trained from scratch, and every test read of
every registered configuration on these 14 datasets is reported.
\item \textbf{An analysis} of which reads carry the gains, from the validation screens that selected each model, and
of cost: thousands of parameters, one CPU thread, and a 1.2-million-entity knowledge graph on one GPU in 18 minutes per
seed (Section~\ref{sec:analysis}).
\end{itemize}

\section{Related work}
\label{sec:related}

\paragraph{Memory and neighbourhood models.}
JODIE~\citep{kumar2019jodie}, DyRep~\citep{trivedi2019dyrep} and TGN~\citep{rossi2020tgn} keep a learned dense memory
per node, updated by message functions. TGAT~\citep{xu2020tgat}, CAWN~\citep{wang2021cawn}, GraphMixer~\citep{cong2023graphmixer},
DyGFormer~\citep{yu2023dygformer}, NAT~\citep{luo2022nat}, TNCN~\citep{zhang2024tncn} and
HyperEvent~\citep{gao2025hyperevent} encode sampled temporal neighbourhoods per query with a deep network.
TPNet~\citep{lu2024tpnet} encodes time-decayed temporal-walk matrices through random projections. It is the closest of
these to our pair records, and it leads \texttt{tgbl-coin-v2} and \texttt{tgbl-flight-v2}. Our records
are numerical and exact rather than learned or projected. The learned part is only the small map that scores what the
records say.

\paragraph{Heuristics and simple baselines.}
EdgeBank~\citep{poursafaei2022edgebank} predicts that seen edges recur. Persistent forecast and moving average are TGB's
node-affinity baselines~\citep{huang2023tgb}. \citet{cornell2025heuristics} show that recency and popularity heuristics
are competitive on TGB link prediction. TRACE~\citep{wu2026trace} combines hand-built recency, decayed-count, popularity,
transition and co-occurrence features with a listwise linear combiner of at most 45 weights. It leads
\texttt{tgbl-wiki-v2} under per-timestamp memory updates. ClockRace shares the premise that well-specified temporal
evidence carries most of the signal. It adds addressed records that make that evidence exact and local, nonlinear maps
that compose many reads, a race among causes, and training races matched to the evaluation. Classical link-prediction
heuristics~\citep{libennowell2007link} motivate our two-hop reads.

\paragraph{Node affinity.}
NAViS~\citep{feldman2026navis} leads all four TGB node-affinity leaderboards. It builds a virtual-state model on the
equivalence between affinity heuristics and state-space models and trains it with a loss designed for affinity
prediction. Our affinity model reads previous-period vectors (persistent forecast and moving average are special
cases) together with affinities at several elapsed-time decay scales. On genre and reddit it adds interaction-count
scales, and a learned per-source softmax over the scales mixes them, every scale contributing. A nonlinear map can
extrapolate trends.

\paragraph{Temporal knowledge graphs.}
RE-GCN~\citep{li2021regcn} and CEN~\citep{li2022cen} evolve entity embeddings over snapshots. TLogic~\citep{liu2022tlogic}
and CountTRuCoLa~\citep{gastinger2026counttrucola} learn temporal rules, and the recurrency baseline predicts that facts
recur~\citep{gastinger2024recurrency}. Our causes correspond to rule families: recurrence; relation transfer, a rule
across relations; transitions; popularity; and two-hop paths. Each cause's confidence is replaced by a hazard learned from
exact record clocks, and all causes compete in one normalized race trained end to end.

\paragraph{Point processes and races.}
Decayed counts are Hawkes-style excitations~\citep{hawkes1971spectra}. The race readout uses a classical fact: among
independent exponential clocks with rates $\lambda_j$, clock $c$ fires first with probability
$\lambda_c/\sum_j\lambda_j$, Luce's choice rule~\citep{yellott1977luce}.

\section{Method}
\label{sec:method}

\subsection{Setting}
A temporal graph is a stream of interactions $(s_i,d_i,t_i,w_i,r_i)$: a source, a destination, a time, an optional
weight and an optional relation type. A \emph{query} $q=(s,t[,r])$ asks for a ranking of a candidate set $C(q)$ of
destinations, or of destination classes for node affinity. The rank of the true destination $y$ is scored by mean
reciprocal rank (MRR) for link tasks. For node affinity, a query at label time $t$ ranks classes for the distribution of
the source's interactions in the following period, scored by NDCG@10. The benchmark fixes $C(q)$: the true destination
plus negatives from a declared generator, or all entities.

\subsection{Addressed time-referenced records}
Let $\mathcal{A}$ be a set of addresses, such as pairs $(s,d)$, destinations $d$, sources $s$, triples $(s,r,d)$ and
pairs $(r,d)$. A record $R_a=(h_a,\tau_a)$ holds numerical history values $h_a\in\mathbb{C}^m$ and a time reference
$\tau_a$. A record exists only once an interaction has addressed it. \emph{Transport} to time $t$ is
\begin{equation}
  T_{t-\tau}\,h = \operatorname{diag}\!\big(e^{(-\rho_k+i\omega_k)(t-\tau)}\big)_{k=1}^m\,h,
\end{equation}
with decay rates $\rho_k>0$ (time constants from a minute to a year; 1 to 8 periods on trade and 1 to 256 time steps on
the knowledge graphs) and optional rotation frequencies $\omega_k$ (daily and weekly cycles). An interaction at time $t_i$ \emph{updates} every record $a$ it addresses: $h_a\leftarrow
T_{t_i-\tau_a}h_a+w_i\mathbf{1}$ and $\tau_a\leftarrow t_i$. No other record changes. Real-coordinate pairs implement
the rotations; a decayed count is the special case $\omega_k=0$. A record can also keep its count, last time and last
counterparty. An \emph{interaction-count} scale weights the $n$-th of a source's $N$ interactions by
$e^{-(N-n)/\kappa}$ in place of elapsed time. The records are numerical state, not parameters, and no gradient flows
through them.

\paragraph{Causal reads.}
A query at time $t$ reads only interactions with $t_i<t$. Interactions that share the query's timestamp, including
those of the same batch, remain pending until a later time is queried. Validation and test interactions update the
records only after the queries at their own time have been scored, as the TGB rules allow for memory
modules~\citep{huang2023tgb}. For batch evaluation, each record family is kept as a time-sorted list of interactions
with the history values after each one. A query finds its record's last interaction before $t$ by one binary search and
transports it to $t$, which reproduces the state of an interaction-by-interaction memory exactly.

\subsection{Causes and the race}
A \emph{cause} $k$ proposes a set of candidates $P_k(q)\subseteq C(q)$ and gives each proposed candidate a log-hazard
$\log\lambda_k(q,c)=g_k(\phi_k(q,c))$. Here $\phi_k$ concatenates the cause's record reads, for example the pair
record $(s,c)$ transported to $t$, with its count, age and seen flag. The map $g_k$ is a small learned network (layer
normalization and a two-hidden-layer GELU network of width 32 or 64; Appendix~\ref{app:models} gives each model's
width) for every candidate-specific cause. On knowledge graphs the popularity and background causes, which
all queries of a (relation, time step) group share, are log-linear in their reads with relation-specific weights.
A candidate's hazard is the
sum of the hazards of the causes that propose it, $\lambda(q,c)=\sum_{k:\,c\in P_k(q)}\lambda_k(q,c)$, and
\begin{equation}
  P(c\mid q)=\frac{\lambda(q,c)}{Z(q)},\qquad Z(q)=\sum_{c'\in C(q)}\lambda(q,c'),
\end{equation}
the probability that $c$'s clock fires first among independent exponential clocks~\citep{yellott1977luce}. With one
cause that proposes every candidate, this is a softmax over $a_c=g(\phi(q,c))$.

\paragraph{Exact races over all entities.}
On knowledge graphs every entity is a candidate. The background cause $\mathrm{BG}$ gives every entity a hazard from
identity-free features: its activity clocks and the set of relations it has taken part in. It is computed once per
(relation, time step) group, as is the popularity cause $\mathrm{POP}$ over the objects the relation has had. The other
causes are sparse. Recurrence ($\mathrm{REC}$) reads the record $(s,r,e)$. Relation transfer ($\mathrm{TR}$) reads
$(s,r',e)$ for $r'\neq r$. Transitions ($\mathrm{TRANS}$) read records $(r,A\to e)$ for the objects $A$ of the subject's
last time step. Two-hop paths ($\mathrm{PATH}$) go from the subject to its recent partners and on to theirs. Since a
sparse cause adds hazard only to its proposals,
$\log Z = \operatorname{logsumexp}_{e}\big(\log(\lambda_{\mathrm{BG}}(e)+\lambda_{\mathrm{POP}}(e)+\sum_{k\,\text{sparse}}
\lambda_k(e))\big)$ is computed exactly over all $N$ entities at the cost of the dense background plus the proposals.
The gradient of $-\log P(y\mid q)$ gives each cause of a losing candidate $c$ its share of the normalizer,
$\lambda_k(c)/Z$. It gives each cause of the true object $y$ that share minus its responsibility
$\lambda_k(y)/\lambda(y)$. Every cause and the background receive counterfactual credit from every query.

\subsection{Evaluation-shaped training races}
\label{sec:races}
The benchmark ranks the true destination against negatives from a declared generator, and we train on races whose
competitors follow that generator:
\begin{itemize}[leftmargin=1.2em,itemsep=1pt,topsep=2pt]
\item \textbf{All candidates.} The race runs over every destination on \texttt{tgbl-wiki-v2} (1,000) and over all
entities on \texttt{tkgl-polecat} and \texttt{tkgl-icews}. The other true objects of the same (subject, relation,
time) are removed from $Z$, as the time-aware filtered evaluation removes them~\citep{gastinger2024tgb2}.
\item \textbf{Historical-random.} On \texttt{tgbl-review-v2}, \texttt{tgbl-flight-v2} and \texttt{tgbl-coin-v2}, TGB's
negatives mix the source's own past destinations with random ones at ratio 0.5~\citep{huang2023tgb}. The
training race is the positive plus up to $\lfloor K/2\rfloor$ competitors from the source's destinations seen before
the query, with the rest uniform over destinations outside that history ($K=20$).
\item \textbf{Node-type uniform.} TGB~2.0 draws heterogeneous-graph negatives uniformly from the true destination's node
type, and so do we ($K=20$ on \texttt{thgl-github} and \texttt{thgl-myket}, $K=200$ on \texttt{thgl-forum}).
\item \textbf{Relation pool.} \texttt{tkgl-wikidata} lists 1,000 candidates per query. The model reads every cause per
listed candidate. It trains on the true object plus 1,000 negatives drawn from the objects its relation has in the
training data, time-filtered like the official generator.
\end{itemize}
The node-affinity model minimizes cross-entropy against the realized class distribution of each label time. That
distribution becomes a training target only after the period it summarizes has closed. All models use AdamW with
learning rate 0.003 and keep the epoch with the best validation score. Training passes over the training stream in time
order, and every read is causal.

\subsection{Models and complexity}
\label{sec:models}
\textbf{Affinity} (\texttt{tgbn-*}). For each (source, class), the model reads previous-period vectors (four
completed periods on trade; on the streaming datasets, the three most recently revealed label vectors and decayed
label averages). It also reads affinities at elapsed-time decay scales, the class's global share and growth, and the
source's volume, growth and recency. Trade adds the reverse flow and pair growth. Genre and reddit add a 120-day scale
and interaction-count scales of 256 and 1,024 interactions. A per-source softmax over the bank of seven scales mixes
them, with zero-initialized bids from the source's context, so the mixture starts uniform.
\textbf{Link} (\texttt{tgbl-*}). The model reads pair, destination and source records. On wiki, the pair records
carry decays from a minute to a month and daily and weekly rotations. The wiki reads add the candidate's recency rank
among the source's destinations, the pair's share of the source's activity, a last-destination flag, transitions from
the source's previous destination, and a decayed co-visitation table. Review adds a popularity residual. Flight
uses day-scale clocks, delayed reads of the pair's activity exactly 1, 7 and 14 days earlier, and the reverse pair.
Coin adds two-hop counterparty reads, per-source and global transitions, flow records, amounts and novelty counts.
\textbf{Heterogeneous} (\texttt{thgl-*}). Relation-addressed families $(s,r,c)$, $(r,c)$ and $(s,r)$ sit beside
$(s,c)$ and $c$, with clocks from one hour to 30 days and an 8-dimensional relation embedding. Forum adds the reverse
pair and source clocks.
\textbf{Knowledge graph} (\texttt{tkgl-*}). This model has the causes REC, TR, TRANS, POP and BG, with clocks of 1 to
256 time steps. On polecat and icews, recency windows restrict each cause to records active in the last 365/30/30/30
steps. Polecat adds
PATH, with at most $16\times16$ paths per query within 30 days; its output layer starts at weight 0 and bias $-8$, so
the cause earns its hazard by gradient.

No model has node embeddings. The wiki model trains one bias per destination (1,000 parameters), and no other model
trains or reads per-entity parameters. An update costs $O(m)$ per
addressed record, independent of the number of nodes. A query costs one read and one map evaluation per candidate and
cause. On knowledge graphs the dense background is shared by all queries of a (relation, time step) group. On
\texttt{tkgl-polecat} the path cause reads about 92 two-hop paths per query; a smoke run measured about 0.4\,ms per
query on one CPU thread.
Appendix~\ref{app:models} lists every configuration and the searched alternatives.

\section{Experimental protocol}
\label{sec:protocol}
\textbf{Data and evaluation.} We use the official splits and negative samples of \texttt{py-tgb} 2.3.0, which meets
every listed package requirement. Node-affinity scores come from the official \texttt{Evaluator}. Link and
heterogeneous-graph scores are computed locally and checked against the official \texttt{Evaluator} on the first 50
queries of each evaluation. Knowledge-graph scores rank the true object against all entities (against the official
listed negatives on wikidata) with the official time-aware filtering and tie rule. Under TGB's batch-of-200 memory
convention our scorer reproduces EdgeBank's posted validation scores within 0.001 (Appendix~\ref{app:scorer}). No external data or pretrained
weights are read.

\textbf{Information boundary.} Each query reads only interactions strictly earlier than its own timestamp (ties
excluded; on knowledge graphs, earlier snapshots). Validation and test interactions update the records after the queries
they answer are scored and never produce a gradient. Validation selects the epoch and the configuration, and each seed's
test score is read once with its best validation checkpoint. On trade, period aggregates are tabulated once from the
event stream, and a forecast at a label time reads only periods that ended at or before it. On wiki, the source and
destination index is allocated over every node ID of the dataset and the official validation negatives, whose union is
all 1,000 destinations.

\textbf{Update granularity.} Our reads see every interaction strictly before the query time, including earlier
interactions of the same batch of 200. TPNet and DyGFormer see the same history through DyGLib's neighbour sampler,
and the wiki entries of TRACE and the LocalGlobal heuristic are evaluated the same way~\citep{wu2026trace}. TGB's
example scripts for memory models update memory once per batch of 200. On \texttt{tgbl-wiki-v2} that convention costs the
heuristics evaluated by \citet{wu2026trace} about 0.08 MRR. Our wiki result is comparable with the per-timestamp
entries, including the leader.

\textbf{Selection, registration and seeds.} Hyperparameters and the best epoch are selected on validation only. Before
the first test read of each configuration, its exact configuration and seeds were fixed in a written registration.
Each seed was then trained from scratch and its test score was read once with the selected checkpoint. Twelve
entries registered and ran seeds 0--4 together. For \texttt{tgbn-trade} and \texttt{tgbl-wiki-v2}, seeds 3 and 4 were
run on 30 October 2026 with the configuration frozen, after the test scores of seeds 0--2 (9 October) were known. An
earlier configuration of \texttt{tgbn-genre}, \texttt{tgbn-reddit} and \texttt{tkgl-polecat} was scored on test with five
seeds (1, 2 and 3 November) before the reported configuration was developed and selected on validation; both are
reported (Table~\ref{tab:earlier}). The polecat path cause was motivated by the first configuration's test error
profile, in which 42\% of test queries ask for an object new to their (subject, relation) pair; the cause was then
built, screened and selected on validation; for this reason we withdrew the polecat entry from the leaderboard on 30
November 2026 (Section~\ref{sec:published}). The reported review and polecat checkpoints were also scored once in TGB's
live-update setting~\citep[Appendix~C]{huang2023tgb}, in which weights adapt after each scored batch (review on 3
November: 0.5440; polecat on 20 November: 0.2588). Those scores are not part of the entries.
Appendix~\ref{app:history} lists every test read.

\textbf{Comparison with the leaders.} We compare with the best posted test score of each leaderboard, read on \leaderboardread{}. Where a
leader posts a standard deviation, we test our five seeds against it with Welch's one-sided $t$-test, assuming five
leader runs as the TGB rules require. Otherwise, we test against the posted mean with a one-sample $t$-test.

\textbf{Hardware.} Every CPU run uses one thread (\texttt{torch.set\_num\_threads(1)}). Ten entries and seeds 3--4
of trade and wiki ran on host A, an AMD Ryzen~9 7950X3D. Seeds 0--2 of trade and wiki ran on host B, an Intel Core
i5-4690, and polecat ran on host C, an Intel Core Ultra~7 265. \texttt{tkgl-wikidata} ran on host C's NVIDIA RTX PRO 6000 Blackwell Max-Q GPU with two CPU threads, using at most
10.8\,GB of GPU memory.

\section{Results}
\label{sec:results}

\begin{table}[t]
\centering
\caption{Test results on 14 TGB and TGB~2.0 datasets: five seeds, mean $\pm$ unbiased standard deviation, higher is
better. The leader is the best posted test score on \leaderboardread{}: NAViS~\citep{feldman2026navis},
TRACE~\citep{wu2026trace}, GraphMixer~\citep{cong2023graphmixer}, TPNet~\citep{lu2024tpnet},
EdgeBank$_\infty$~\citep{poursafaei2022edgebank}, TGN with edge types~\citep{rossi2020tgn,gastinger2024tgb2} and
CountTRuCoLa~\citep{gastinger2026counttrucola}. $\Delta$ is our mean minus the leader's; ``$>$\,leader'' counts the
seeds above it. Every
difference is significant at $p<0.05$ (one-sided; Section~\ref{sec:protocol}), and all but review at $p<0.005$.
Parameters are counted as TGB counts them, \texttt{sum(p.numel())}. Time is the median wall-clock minutes per seed,
including training, validation and the test pass.}
\label{tab:main}
\scriptsize
\setlength{\tabcolsep}{3.2pt}
\begin{tabular}{@{}llrrrrcrr@{}}
\toprule
Dataset & Leader & Score & Ours, validation & Ours, test & $\Delta$ & $>$\,leader & Params & Min. \\
\midrule
\mainrows
\bottomrule
\end{tabular}
\end{table}

Table~\ref{tab:main} gives the main results, and Appendix~\ref{app:seeds} lists every seed. On every one of the 14 datasets the mean test score exceeds the best posted
test score, and so does every seed. The metric is NDCG@10 for \texttt{tgbn-*} and MRR otherwise.
Section~\ref{sec:published} compares each dataset with the strongest published results outside the leaderboards.

\textbf{Largest margins.} On \texttt{thgl-github} and \texttt{thgl-myket} the only posted entries are EdgeBank variants.
TGB~2.0 reports that most neural methods it evaluated ran out of memory or time on its larger heterogeneous
graphs~\citep{gastinger2024tgb2}. ClockRace
doubles their MRR (0.847 against 0.413 and 0.916 against 0.456) with about 7,000 parameters on one CPU thread.
Outside the leaderboard, TSNN~\citep{anon2026tsnn} scores 0.834 on github, below ours, and 0.922 on myket, above ours
(Section~\ref{sec:published}). On
\texttt{tkgl-icews} it reaches 0.345 against 0.214, on \texttt{thgl-forum} 0.816 against 0.729, and on
\texttt{tkgl-wikidata} 0.671 against 0.609. RE-GCN and CEN ran out of memory on wikidata in TGB~2.0, and the recurrency
baseline ran out of time.

\textbf{Smallest margins.} The margin is $+0.0021$ on \texttt{tkgl-polecat}, $+0.0023$ on \texttt{tgbl-wiki-v2}
against TRACE, $+0.0026$ on \texttt{tgbn-token} and $+0.0057$ on \texttt{tgbn-trade}. On these four datasets the
leaders read closely related evidence: rules over recurrent facts, recency and co-occurrence features, and state-space
forecasts of past affinity vectors. The seed spread of our models is 0.0002 to 0.0013, so each of these margins
still spans several standard deviations. \smallmarginsnote

\begin{table}[t]
\centering
\caption{Earlier registered configurations of the same datasets, read on test before the reported ones (validation
and test: five seeds, mean $\pm$ SD). The last column repeats the reported configuration's test mean from
Table~\ref{tab:main}. The polecat configuration without the path cause was above the leader of its time, TLogic at
0.228. It is below CountTRuCoLa, posted on 13 November 2026.}
\label{tab:earlier}
\scriptsize
\begin{tabular}{@{}llrrcr@{}}
\toprule
Dataset & Earlier configuration & Validation & Test & Seeds $>$ leader & Reported test \\
\midrule
\earlierrows
\bottomrule
\end{tabular}
\end{table}

\section{Published results outside the leaderboards}
\label{sec:published}

A leaderboard ranks only the entries posted on it. Table~\ref{tab:published} compares each dataset with the strongest
published result we found, including papers that never posted to the leaderboard, among methods that use the same or
less information than our models. Information here is what the benchmark reveals by the query time: an earlier event's
own features count, such as a review's star rating, even where our model does not read them; a quantity collected after
the event, such as a reply's score at crawl time, does not. Higher printed scores that use more information, or that do
not hold or are not established under TGB's evaluation, are listed beside it, and the paragraphs below give the
reasons. The literature search and the reading of the comparators' released code were done on 30 November and
1 December 2026, and on 3 December 2026 the search was extended to the submissions and public discussions on
OpenReview.

\begin{table}[t]
\centering
\caption{The strongest published result with the same or less information, per dataset, as printed in its paper
($\pm$ where printed): \publishedbarcites. Ours: test mean over five seeds.
$\Delta$ is our mean minus that result; the next column counts our seeds above it. The last column lists higher printed
results that are not comparable: $^a$the printed run read the edges of the label year, as its authors confirmed (their
strict-cutoff rerun: 0.862); $^b$label-dependent test path (TGR~\citep{petrovic2024tgr}); $^c$TGB's test negatives
shifted by one node ID (CRAFT~\citep{yi2025craft}); $^d$evaluation not established as TGB's
(A2QTGN~\citep{innan2026a2qtgn}).\unresolvednote{}\prismcaptionnote{} $^h$Reads the forum's event features, which carry
each reply's score at crawl time (TSNN). $^i$Updates its memory after every query, so a query reads true facts of its own
time step (RF-Logic~\citep{anon2026rflogic}). The
leakage-free rows use the protocol of \citet{wang2026leakage}
against its best published score, their hybrid blend of training-free statistics with weights fit on validation.}
\label{tab:published}
\scriptsize
\setlength{\tabcolsep}{3.5pt}
\begin{tabular}{@{}lrlrrcl@{}}
\toprule
Dataset & Ours & Strongest published & Score & $\Delta$ & Seeds $>$ & Higher, not comparable \\
\midrule
\publishedrows
\midrule
\leakagefreerows
\bottomrule
\end{tabular}
\end{table}

ClockRace leads the strongest such result on \publishedleadcount{} of the 14 datasets with every seed, by
\publishedleadmin{} to \publishedleadmax{}, and on \texttt{tgbn-reddit} and \texttt{tgbn-token} under the leakage-free
protocol. It trails on \publishedtrails{} and, under TGB's official evaluation loop, on reddit and token\levelclause{}\unresolvedclause{}.

On the heterogeneous and knowledge graphs, two results published outside the leaderboards narrow the margins over the
posted entries. TSNN~\citep{anon2026tsnn} keeps each node's memory in its own coordinate frame and, with TGB's negatives
and evaluator, scores all queries of a timestamp from the events before it. It is the strongest such result on
\texttt{thgl-github}, \texttt{tkgl-icews} and \texttt{tkgl-wikidata}, where ClockRace leads it by $+0.0130$, $+0.0581$
and $+0.0205$ with every seed, and on \texttt{thgl-myket}, where its 0.9219 is above our 0.9158; on github it narrows
our margin over the posted entries to $+0.0130$ from $+0.43$. On the link datasets it prints 0.8290 on wiki, 0.9040 on
flight and 0.8294 on coin, below the results in Table~\ref{tab:published}, and 0.5400 on review, discussed below.
EdgeReMIND~\citep{pollard2026edgeremind}, a memorization baseline, is below ClockRace on every dataset both report in
its main results. Its paper states that a query reads only earlier edges; its released code scores the test stream in blocks of
200 events, and every query of a block reads the events before the block's start, so earlier events of its own block are
not read and events of earlier blocks at the query's timestamp are (\texttt{thgl/edgeremind.py} lines 137 and
207--208 and \texttt{tkgl/edgeremind.py} lines 125 and 195--196 under \texttt{examples/linkproppred}, and
\texttt{tgm/nn/modules/edgeremind\_benchmark.py} line 341, at commit \texttt{41c0db2}). Our reads see every strictly
earlier interaction (Section~\ref{sec:protocol}).

\textbf{Where ClockRace trails.} On \texttt{thgl-myket}, TSNN's 0.9219 is above our 0.9158 and above every one of our
seeds (highest 0.9164). On \texttt{tkgl-polecat}, CognTKE~\citep{chen2025cogntke}, run with its authors' code
by the authors of the leading method, scores 0.287 in Table~1 of their paper~\citep{gastinger2026counttrucola}, above our
0.2581; RF-Logic prints 0.424 with the per-query memory updates described below. Our polecat configuration's path cause was
motivated by the first configuration's test error profile (Section~\ref{sec:protocol}); for that reason we withdrew its
leaderboard entry on 30 November 2026, and we report it here with that history. On \texttt{tgbn-reddit} and
\texttt{tgbn-token}, DyNA-Mixer~\citep{zheng2026dynamixer} scores 0.590 and 0.523 under TGB's official loop, above our
0.5868 and 0.5156; both methods read the labels that this loop returns early (Section~\ref{sec:limitations}), as
DyNA-Mixer's authors confirmed for its genre, reddit and token scores (personal communication, 3 December 2026). Under the
leakage-free protocol of \citet{wang2026leakage}, ClockRace exceeds the best published score on genre, reddit and token
with every seed, and on genre it also leads DyNA-Mixer's 0.544 under the official loop.

\textbf{Forum: event features and per-query updates.} On \texttt{thgl-forum}, TGB's event features are each reply's word
count and its Reddit score, each divided by its maximum. The score is the one at crawl time, which includes votes cast
after the reply and after later query times. TSNN's state updates read every event's features (its Section~2.1 and
Eq.~(4)), and its supplementary code sets their width from the dataset for every TGB dataset (\texttt{scripts/train.py}
lines 396--397), so its 0.8381 reads these scores; TGN with edge types, the posted leader at 0.729, reads them as
message features. RF-Logic~\citep{anon2026rflogic} prints 0.739 on forum and 0.424 on polecat. A public comment on its
OpenReview page by an author of TGB~2.0 reports that, in its supplementary code, \texttt{test()} appears to update the
memory after every query, so that later queries of a time step are scored with that step's true facts, on knowledge
graphs including each query's own inverse fact; its authors had not answered by 3 December 2026. Our forum model reads
neither the event features nor any event at the query's own time, so we compare it with EdgeReMIND's 0.7306, which it
exceeds by $+0.0856$ with every seed; it is also above RF-Logic's printed 0.739.

\prismparagraph

\textbf{Scores made with TGB's test negatives shifted by one node ID.} CRAFT~\citep{yi2025craft} prints 0.9139 on
\texttt{tgbl-flight-v2} and 0.8847 on \texttt{tgbl-coin-v2} (CRAFT-R). Its released TGB loader looks up each test edge's
negatives in TGB's node IDs and then adds 1 to every source, destination and edge ID, but not to the negatives
(\texttt{utils/DataLoader.py}, lines 176--200 at commit \texttt{81590a7}), so each negative $n$ is scored as node $n-1$.
TGB draws up to half of these negatives from the source's earlier destinations; after the shift most of them are other
nodes. Pull request~\#2 on the authors' repository\footnote{\url{https://github.com/luyi256/CRAFT/pull/2}} (opened 20
October 2026, not merged) adds the missing shift and
reports, for the published configuration, 0.865 on flight (one run) and 0.747 on coin (three runs). MambaTG
\citep{nguyen2026mambatg} reruns CRAFT-R under TGB's official negatives at 0.864 and 0.744. CL-DGFM's 0.9068 on flight
and CTDG-SSM's 0.862 on coin state TGB's evaluation, and ClockRace exceeds both with every seed; CTDG-SSM updates its
memory once per batch of 128 events (its Section 5 and Table 9), while our reads see every strictly earlier interaction.

\textbf{A test path that depends on the labels.} TGR~\citep{petrovic2024tgr} prints 0.854 (TGR-TGN) and 0.599
(TGR-TNCN) on \texttt{tgbl-review-v2}. In its released \texttt{test()} (lines 126--150 of \texttt{tgr-tgn-b.py} and
\texttt{tgr-tgn-bn.py} and lines 168--194 of \texttt{tgr-tncn-b.py} and \texttt{tgr-tncn-bn.py}, at commit
\texttt{3709930}), a candidate takes the expander embedding when it is an
endpoint of a positive edge of the batch being scored and was observed in an earlier batch; in \texttt{train()}, every
previously observed node of the batch takes it, sampled negatives included. Issue~\#1 on the authors'
repository\footnote{\url{https://github.com/kpetrovicc/TGR/issues/1}} (opened 21 November 2024) measured on the review test split that 91.98\% of true destinations take this path against 6.51\% of
negatives; the authors replied on 16 December 2024 that they were assessing the test loop, and the repository has no
later commit. \reviewsentence{} A2QTGN~\citep{innan2026a2qtgn} prints 0.5883 on review, without spread, run count or
code. Its first version scores against its own negatives, items the user has not interacted with; its second version
names TGB's negatives, with every table value unchanged. We list it in the last column of Table~\ref{tab:published}
rather than as a result under TGB's evaluation.

\textbf{Edges of the label year.} DyNA-Mixer prints 0.869 on \texttt{tgbn-trade}. TGB builds the trade label of year $Y$
from the normalized trade flows of year $Y$, and TGB's evaluation loop passes every edge stamped $Y$ to the model before
it predicts that label (\texttt{examples/nodeproppred/tgbn-trade/tgn.py}, \texttt{test()}, lines 241--246 at TGB commit
\texttt{0e12151}). DyNA-Mixer's Section~3.1 defines the graph observed at time $t$ as every event with $\tau\le t$, and
its Section~4 takes that stream as the input at the label time; its Section~1 describes the forecast as conditioned on
all observations before $T_q$. Its authors have confirmed that the run behind 0.869 followed the order of TGB's
node-property examples, so that almost all edges of year $Y$ reached the model before it predicted that year's label,
and that their paper will be corrected. Their rerun with a strict cutoff, which holds back every edge at or after the
next unreleased label time until that label has been predicted, reads $0.8620\pm0.0009$ over five seeds, and
$0.8624\pm0.0009$ without the graph module (personal communication, 3 December 2026). Under TGB's loop a parameter-free
edge counter scores 1.000; with a strict cutoff the best published score is 0.863, reached by NAViS in both of its modes
and by two blends~\citep{wang2026leakage}. Our trade model reads only earlier years' edges and leads NAViS and the
strict-cutoff DyNA-Mixer with every seed.

\section{Analysis}
\label{sec:analysis}

Table~\ref{tab:screens} collects the validation measurements that selected each model, with training-free scorers
evaluated under the same causal reads. They show where the gains come from.

\begin{table}[t]
\centering
\caption{Validation evidence (seed 0 unless noted; MRR or NDCG@10). Training-free scorers use the same strictly causal
reads as the models. All values are validation scores; the polecat path-cause, genre and reddit rows were measured
after the earlier configurations' test reads (Table~\ref{tab:earlier}).}
\label{tab:screens}
\scriptsize
\setlength{\tabcolsep}{4pt}
\begin{tabular}{@{}lll@{}}
\toprule
Dataset & Comparison & Validation \\
\midrule
\texttt{thgl-github} & relation-addressed records / untyped records / causal seen-pair record (EdgeBank) & \textbf{0.7558} / 0.7337 / 0.464 \\
\texttt{thgl-myket} & relation-addressed records / untyped records & \textbf{0.9165} / 0.9127 \\
\texttt{thgl-forum} & with / without reverse-pair read and source clocks ($K=100$, two epochs) & \textbf{0.8163} / 0.7716 \\
\texttt{tkgl-polecat} & with / without two-hop path cause; on queries whose object is new to $(s,r)$ & \textbf{0.2635} / 0.2597; 0.083 / 0.077 \\
\texttt{tgbn-genre} & race over seven decay scales / single bank (v2) & \textbf{0.5385} / 0.5365 \\
\texttt{tgbn-reddit} & race over seven decay scales / single bank (v2), five-seed means & \textbf{0.6008} / 0.5981 \\
\texttt{tgbl-review-v2} & trained race / popularity minus seen pair / 30-day popularity & \textbf{0.4880} / 0.479 / 0.341 \\
\texttt{tgbl-flight-v2} & trained race (five seeds) / recency then popularity & \textbf{0.9147} / 0.8757 \\
\texttt{tgbl-coin-v2} & trained race (five seeds) / pair recency then popularity & \textbf{0.8536} / 0.7813 \\
\texttt{tgbl-coin-v2} & pair records + paths, transitions, flow, amounts, novelty / pair records only & \textbf{0.8537} / 0.8411 \\
\texttt{tkgl-polecat} & with / without the relation-transfer cause (first configuration) & \textbf{0.2597} / 0.2214 \\
\texttt{tgbn-trade} & trained race (five seeds) / persistent forecast & \textbf{0.8746} / 0.8604 \\
\texttt{tgbl-wiki-v2} & without / with learned source--destination identity embeddings & \textbf{0.8518} / 0.8515 \\
\bottomrule
\end{tabular}
\end{table}

\textbf{Addressing is the main lever.} On \texttt{thgl-github}, addressing records by relation type lifts validation
MRR from 0.734 to 0.756, and both far exceed a causal record of seen pairs (0.464). On \texttt{thgl-forum} the reverse
pair, read at the swapped address of the same record family, together with source clocks, adds 0.045: replies are
reciprocal. On \texttt{tkgl-polecat}, 42\% of test queries ask for an object that the (subject, relation) pair has
never had. Two-hop paths reach the true object of 82\% of these queries on validation, and the path cause lifts the
whole validation score by 0.0038.

\textbf{The training race must contain the evaluation's competitors.} On \texttt{tgbl-review-v2} up to 50 of each
query's 100 negatives are the reviewer's own past items, and a reviewer reviews an item again in 0.17\% of validation
queries. Models trained on random competitors learn a seen pair as positive evidence and rank those negatives first.
Rule-based checks show the size of the effect: subtracting a large constant for seen pairs lifts 30-day popularity from
0.341 to 0.479. The trained race, whose competitors include the source's own history, learns this satiation from data
and reaches 0.488. The posted leader, GraphMixer, has validation MRR 0.428. The same principle selects node-type-uniform
competitors on heterogeneous graphs and relation-pool competitors on \texttt{tkgl-wikidata}.

\textbf{Learned races improve on the strongest causal heuristics we built.} With exactly causal reads,
recency-then-popularity rankings are strong: 0.876 on flight and 0.781 on coin. Our strongest causal heuristics reach
0.888 on flight (the pair's count on the same day one week earlier, then recency) and 0.794 on coin (a one-day pair
clock, then popularity). The posted coin heuristic, LocalRecencyLocalPopularity, scores 0.774 on
test~\citep{cornell2025heuristics}. On seed-0 validation, the selected races add 0.027 and 0.060 to our strongest
heuristics. Beyond pair records,
the coin model reads two-hop counterparty paths, transitions, flow records, amounts and novelty (Appendix~\ref{app:models}).
On trade the race adds 0.014 over the persistent forecast on validation and 0.015 on test (0.8687 against
0.8541, our computation of the persistent forecast; the leaderboard posts 0.855).

\textbf{No identities are needed.} On wiki, learned source--destination embeddings change validation MRR by $-0.0003$,
and the selected model has none; it keeps a learned bias per destination (1,000 of its 7,995 parameters). On review,
identity tables of 11.6 million parameters overfit and were switched off. Of review's 11,642,367 parameters (the sum of
\texttt{model.parameters()}), 11,637,021 are these per-node tables, held at zero, neither trained nor read by the
reported configuration, and 5,346 are trained. Apart from wiki's destination biases, no reported model trains or reads
per-entity parameters, so every model applies unchanged to entities first seen during evaluation.

\textbf{Cost.} Thirteen of the 14 models have 2,039 to 81,178 parameters; review has 11,642,367, of which it trains 5,346. A seed takes 4
minutes on trade, 35 on wiki and 70 on github, on one CPU thread. The longest CPU runs are forum (821 minutes) and
icews (513 minutes, 27\,GiB peak memory). On \texttt{tkgl-wikidata}, with 1.2 million
entities, a seed takes 18 minutes on one GPU with at most 10.8\,GB of allocated GPU memory (Appendix~\ref{app:compute}).

\textbf{Where ClockRace does not lead.} On \texttt{thgl-software} our causal reads reached 0.43--0.47 validation MRR
against STHN's 0.764. The gap sits in entities created during the evaluation period, so we registered no model and read
no test score. Against published results outside the leaderboards, ClockRace trails on \publishedtrails{} and, under
TGB's official loop, on reddit and token\levelclause{}\unresolvedclause{} (Section~\ref{sec:published}).

\section{Limitations and scope}
\label{sec:limitations}
Each model's reads were chosen from validation diagnostics of its dataset family, and the 14 configurations are not
one shared hyperparameter setting. The family shares the record model, the race readout and the training principle.
Test-period interactions update the records after their queries are scored, and parameters are fixed after training,
as the TGB rules require. Three datasets carry a second registered configuration, and both are reported. Results on
\texttt{tgbl-comment} and \texttt{tkgl-smallpedia} are outside this report. Table~\ref{tab:main} compares with posted
leaderboard scores and Table~\ref{tab:published} with published results outside the leaderboards, as their papers print
them; we did not rerun the leaders. Seed spreads were measured on our hosts. Rerun from
the release code (Appendix~\ref{app:repro}) on another host, trade seeds 0, 1, 2 and 4 and wiki seeds 0--2 reproduce bit for bit, and wiki seeds 3--4
agree within $1.1\times10^{-8}$; polecat seed 0, rerun on its own host, reproduces bit for bit. Trade seed 3 moved its best epoch on a flat validation optimum, changing its test score by
$-0.0017$ (1.3 seed standard deviations); with the rerun, the five-seed test mean is 0.8683, still above the leader's
0.863. Ablations on test and a matched-hardware runtime comparison with the leaders are left to future work.

\textbf{Node-affinity labels.} The node-affinity scores use TGB's official evaluation loop, as every leaderboard entry
does. \citet{wang2026leakage} show that this loop is not point-in-time correct: on genre, reddit and token each label
summarises one day about five days after its timestamp, yet the loop returns it to the model right after it is
scored, so a method that reads past labels receives them early. Our trade model reads only edges stamped before each
label time, which is the audit's leakage-free protocol. Our models for the three daily datasets read the returned
labels, as the persistent-forecast and moving-average baselines and the label mode of NAViS~\citep{feldman2026navis} do.
Retrained and scored under the audit's leakage-free protocol (a strict edge cutoff, and each label released only
after its availability time, in training, validation and test alike), they score $0.5330\pm0.0003$ on genre, $0.5663\pm0.0001$ on reddit and $0.4580\pm0.0004$ on token (five seeds each,
every seed above the audit's best), against the audit's best leakage-free scores of 0.530,
0.561 and 0.454 (a hybrid blend of training-free statistics) and NAViS's 0.520, 0.552 and 0.444. Replayed through the same loop, the
persistent forecast reproduces the audit's values on all three datasets.

\section{Conclusion}
Addressed time-referenced records and a race among candidate causes are enough to exceed the posted leaders of 14
temporal graph leaderboards across four task families, and the strongest published results that use the same information
on \publishedleadcount{} of these datasets, with 2,039 to 81,178 trained parameters, one CPU thread per seed and one GPU
for \texttt{tkgl-wikidata}. On these benchmarks the hypothesis we started from holds: a sparse, causal model built like the
processes it observes leads, and updates its state only where events land; on knowledge graphs, the queries of each
relation and time step share one dense background. The results point to a design rule for
temporal graph learning: store evidence where queries will look for it, read it exactly at query time, and train on the
competitors the task will present.

\section*{Use of AI assistance}
The research was carried out with extensive use of AI coding agents built on large language models. The agents were
used for implementation, experiment execution, analysis and drafting. Tables~\ref{tab:main}, \ref{tab:earlier},
\ref{tab:published}, \ref{tab:seeds} and~\ref{tab:compute} are generated from the saved run records by script; every
other result number is taken from those records, leader values from the live leaderboards, and published values from the
cited papers and, for RF-Logic, its authors' public responses. The author takes full responsibility for the content. No
language model is part of the method.

\bibliographystyle{plainnat}
\bibliography{refs}

\appendix

\section{Models, configurations and searched hyperparameters}
\label{app:models}
Selected values are starred; every search used validation only (Section~\ref{sec:protocol}). ``Screened arms'' were
trained on seed 0. Each selection rule was registered before the arms it compares were scored. Unless an entry says
otherwise, the rule takes the arm with the higher best validation score and, within 0.002, the cheaper arm (fewer
parameters, causes or training queries).

\begin{itemize}[leftmargin=1.2em,itemsep=2pt]
\item \textbf{tgbn-trade}: four previous periods; affinities at four decay time constants in periods; reverse flow;
class share and growth; pair growth; node volume, growth and periods since activity (29 features); hidden 32; weight
decay 0.001; up to 200 epochs, best validation epoch kept. No search.
\item \textbf{tgbn-genre}: streaming model, decay scales 1, 3, 7 and 30 days; clock set [single bank, v2 (validation
0.5365); seven-scale race, v3 (0.5385); v3 with the label-time handling of v2, v3.1* (0.5385)]; 4 epochs; hidden 32;
edge weight = official edge feature.
\item \textbf{tgbn-reddit}: as genre; clock set [v2 (0.5978), v3* (0.6010)]; edge weight [count*], set from training
data before any validation score because the official feature (comment score) is negative on 5.3\% of edges.
\item \textbf{tgbn-token}: streaming model without the extra scales, configuration carried over from genre (4 epochs,
hidden 32); edge weight [count, feature*], chosen on training data before any validation score.
\item \textbf{tgbl-wiki-v2}: 41 reads per candidate; source--destination identity dimension [16 (0.8515), 0* (0.8518)];
learned destination bias; 3 epochs; hidden 64; batch 200.
\item \textbf{tgbl-review-v2}: pair, destination and source clocks at 1, 7, 30 and 365 days; popularity residual;
historical-random race with $K=20$ and up to half of the competitors from the source's history; identity tables [on, off*];
explicit refractory term on seen pairs [on (0.48792), off* (0.48802)]; 3 epochs; hidden 64; weight decay $10^{-4}$.
\item \textbf{tgbl-flight-v2}: screened arms [$K=20$ over the full training stream* (0.91476), $K=40$ (0.91432),
$K=20$ over the latest 15 million training queries with 3 epochs (0.91506)]; registered rule: the highest best
validation score and, within 0.002 of it, the arms in the registered order $K=20$ over the full stream, then the
recent-data arm; 2 epochs; hidden 64; 40 reads per candidate.
\item \textbf{tgbl-coin-v2}: read groups over development rounds (seed-0 validation): pair records 0.8411; +paths
0.8449; +transitions 0.8449; +paths and transitions 0.8484; +transitions and flow 0.8486; +transitions and novelty
0.8487; +transitions and amounts 0.8456; all five groups* 0.8537. A width-128 variant with a two-hop sketch (0.8555) was
within the registered 0.002 margin, so the width-64 model was kept; a further round of hazard, period and profile reads
stayed below its registered bar and was not selected. $K=20$; 2 epochs; hidden 64; 128 reads per candidate.
\item \textbf{thgl-github, thgl-myket}: relation-addressed records [on*, off] (0.7558 / 0.7337 and 0.9165 / 0.9127);
$K=20$; 3 epochs; hidden 64; relation embedding 8.
\item \textbf{thgl-forum}: reverse pair and source clocks [on* (0.8163), off (0.7716)] at $K=100$ and 2 epochs, below
the registered gate of 0.817; a second registered screen compared race size $\times$ epochs [$200\times4$* (0.8171),
$300\times3$ (0.8161)], where a difference of at most 0.002 selects $200\times4$; hidden 64.
\item \textbf{tkgl-polecat}: windows REC/TR/TRANS/POP [365/30/30/30 days*]; relation transfer [on* (0.2597), off
(0.2214)] in the first configuration (default rule). A later registered screen took the higher of two path arms [on
(0.26312), on with cause heads of width 64* (0.26351)] against the first configuration rerun on that host (0.25975); at
most $16\times16$ paths in 30 days; 2 epochs.
\item \textbf{tkgl-icews}: none searched; the polecat configuration (windows 365/30/30/30, relation transfer, hidden 32,
2 epochs) was run once on validation (0.3472) as the only candidate.
\item \textbf{tkgl-wikidata}: causes read per listed candidate; training span [all training data (0.7383), from 1950*
(0.7382, 42\% of the work)]; hidden / type dimension / learning-rate schedule [32/16/constant, 64/32/cosine* (0.7406)];
popularity hazard [log-linear*, MLP (0.7377, stopped)]; rule: the higher best validation score and, within 0.001, the
cheaper arm; 1,000 training negatives; 4 epochs.
\end{itemize}

\section{Per-seed results}
\label{app:seeds}
\begin{table}[H]
\centering
\caption{Validation/test score of every seed of the reported configurations.}
\label{tab:seeds}
\scriptsize
\begin{tabular}{@{}lccccc@{}}
\toprule
Dataset & Seed 0 & Seed 1 & Seed 2 & Seed 3 & Seed 4 \\
\midrule
\seedrows
\bottomrule
\end{tabular}
\end{table}

\section{Test-read history}
\label{app:history}
Every test read of a model of ours on these datasets (dates in 2026, UTC; five seeds unless stated):
trade, seeds 0--2 on 9 October and seeds 3--4 on 30 October; wiki, the same dates (EdgeBank and a recency heuristic were
also scored on wiki test on 9 October as baselines); genre, an earlier configuration on 1 November and the reported one
on 3 November; reddit, 2 and 4 November likewise; token, 2 November; review, 2 November, with the same checkpoints
scored in the live-update setting on 3 November; github, 3 November; myket, 3--4 November; forum, 5 November; flight,
5--6 November; icews, 6--7 November; coin, 8 November; wikidata, 6 November (EdgeBank scored in the same jobs);
polecat, an earlier configuration on 3 November and the reported one on 20 November, with its checkpoints scored in the
live-update setting on 20 November. The release code's reproduction runs read test again with the same configurations
and seeds: trade and wiki on 19 November, and polecat seed 0 on 25 November (Section~\ref{sec:limitations}). The
leakage-free runs of Section~\ref{sec:limitations} read test once per seed: genre seeds 0--4 and reddit and token seeds 0--1
on 25 November, and reddit and token seeds 2--4 on 26 November. \cleanroomreads{} The paired check of
Section~\ref{sec:published} read PRISM-TNCN, not a model of ours, on the wiki test set at 21:48 and 22:18 UTC on
1 December.

\section{Scorer validation}
\label{app:scorer}
Under TGB's convention of updating memory once per batch of 200 interactions, our implementation of EdgeBank
reproduces the posted validation MRR of EdgeBank$_\infty$ and of time-window EdgeBank: 0.16600 and 0.36334 against
0.166 and 0.363 on \texttt{tgbl-flight-v2}, and 0.3154 and 0.4915 against 0.315 and 0.492 on \texttt{tgbl-coin-v2}. On
\texttt{tkgl-wikidata} our evaluation path gives EdgeBank a test MRR of 0.5367 ($\infty$) and 0.5369 (time window),
against the posted 0.535. Every link and heterogeneous-graph evaluation checked agreement with the official
\texttt{Evaluator} on its first 50 queries.

\section{Compute resources}
\label{app:compute}
\begin{table}[H]
\centering
\caption{Scale and cost per seed: entities (users + classes for streaming node affinity), test queries, parameters
counted as TGB counts them, peak resident memory, median wall-clock minutes and host. Host A: AMD Ryzen~9 7950X3D;
B: Intel Core i5-4690; C: Intel Core Ultra~7 265 with an NVIDIA RTX PRO 6000 Blackwell Max-Q. Every CPU run uses one
thread.}
\label{tab:compute}
\scriptsize
\begin{tabular}{@{}lrrrrrl@{}}
\toprule
Dataset & Entities & Test queries & Parameters & Peak GiB & Minutes & Host \\
\midrule
\costrows
\bottomrule
\end{tabular}
\end{table}

\section{Reproducibility}
\label{app:repro}
The code of every reported model, at the version that produced these results, is at \codeurl. For each dataset, a
script runs the five seeds with the exact recorded settings, and a results file lists every seed's validation and test
scores. Dependencies are Python, PyTorch, NumPy and \texttt{py-tgb} 2.3.0. The data are downloaded with
\texttt{py-tgb}'s own loader under each dataset's terms; the official negative samples and \texttt{Evaluator} are
used unchanged.

\section{Broader impact}
Temporal interaction prediction supports recommendation, fraud and intrusion detection, capacity planning and
forecasting of political events. The same capability can support profiling of individuals. The models here are small,
read explicit per-entity histories, and are trained only on public benchmark data. Deployment on personal data falls
under the applicable data-protection law and the purpose limits of the data.

\end{document}